\documentclass[letterpaper, 10 pt, conference]{ieeeconf}  

\IEEEoverridecommandlockouts                              

\usepackage[utf8]{inputenc}
\usepackage[T1]{fontenc}
\usepackage{amsmath,amssymb,amsfonts}
\usepackage{graphicx}
\usepackage{booktabs}
\usepackage{xcolor}
\usepackage{hyperref}
\usepackage{cite}
\usepackage{caption}
\usepackage{subcaption}
\usepackage{multirow}
\usepackage{url}
\usepackage{balance}
\usepackage{algorithm}
\usepackage{algorithmic}
\usepackage{float}

\newcommand{\ourmethod}{Gated VLA-Cache}
\hypersetup{colorlinks=true,linkcolor=blue,citecolor=blue,urlcolor=blue}

\begin{document}
\title{\Large \bf Neural Introspection Gating for Adaptive KV-Cache Reuse\\in Vision-Language-Action Models}
\author{Zhijie Wu$^{1}$\thanks{$^{1}$ The authors are with Graduate School of Information Science and Technology, The University of Tokyo, 7-3-1 Hongo, Bunkyo-ku, Tokyo, 113-8656, Japan.[z-wu, kawaharazuka, k-okada]@jsk.imi.i.u-tokyo.ac.jp}, Kento Kawaharazuka$^{1}$, Kei Okada$^{1}$}

\maketitle


\begin{abstract}
Vision-Language-Action (VLA) models map camera images and language
instructions directly to motor commands through a single autoregressive
transformer.
In real-time control, they still spend substantial compute recomputing
key-value (KV) representations for visual tokens that barely change across
neighboring frames.
Recent work such as VLA-Cache reduces that cost by reusing KV states for
visually static patches, but its policy relies only on observation-space
heuristics and does not account for the model's own uncertainty.
We propose \textbf{\ourmethod{}}, a lightweight, training-free extension
that augments visual-similarity caching with \emph{neural introspection}.
The method monitors the \textbf{logit margin} between the top-two predicted
action tokens, a zero-cost confidence signal available during decoding.
When the margin drops below a threshold, the cache is invalidated and a
full recompute is triggered.
Evaluated on four LIBERO benchmark suites with both OpenVLA and
OpenVLA-OFT, \ourmethod{} improves reliability when blind caching hurts.
On LIBERO-Goal and LIBERO-Long, it recovers over 100\% of the lost accuracy
while retaining 80\% of the compute savings.
\end{abstract}

\section{Introduction}\label{sec:intro}

Vision-Language-Action (VLA) models~\cite{kim2024openvla,zitkovich2023rt,black2024pi_0}
use a single autoregressive transformer that ingests a camera frame and a
language instruction, then directly outputs discretized motor commands.
Because they build on large vision-language models (VLMs), VLAs
generalize across manipulation tasks without much task-specific
engineering.
However, running a 7-billion-parameter model at the 10--20\,Hz rates
needed for manipulation is computationally expensive.

Consecutive camera frames in a manipulation episode are largely
redundant.
VLA-Cache~\cite{xu2025vlacache} exploits this by identifying visually
static image patches via pixel-level cosine similarity, reusing
their cached KV representations from the previous forward pass, and
modulating per-layer reuse proportions through an entropy-adaptive
schedule.

The problem is that VLA-Cache's reuse decision is based
entirely on observation-space heuristics (patch similarity and
attention-weighted task relevance).
It cannot tell when the model itself is uncertain,
e.g., during grasp alignment or when a visually small change turns out
to be behaviorally important.
When that happens, stale cached representations inject errors into the
autoregressive decoding of the action vector, and those errors accumulate
over long horizons.

\begin{figure}[t]
  \centering
  \includegraphics[width=\linewidth]{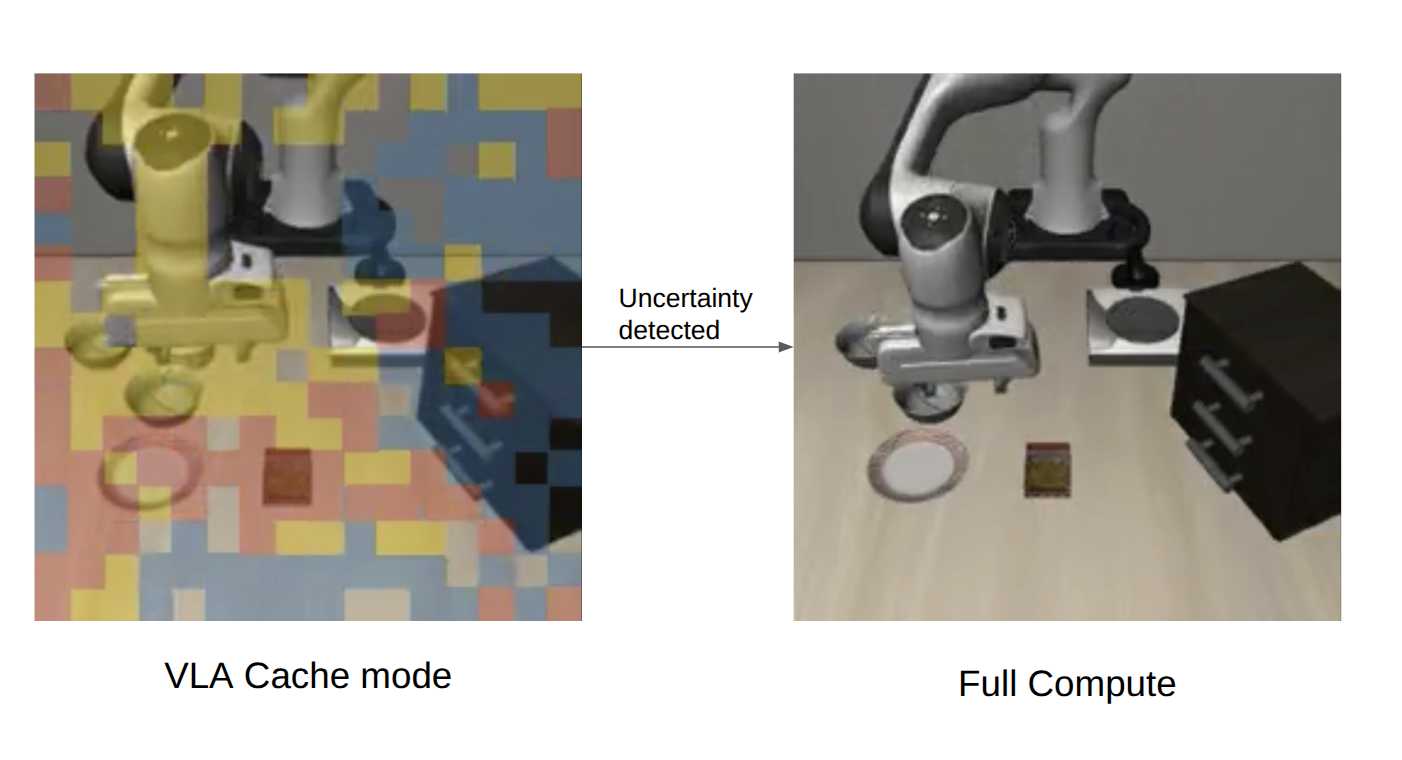}
  \caption{%
    \textbf{Overview.}
    VLA-Cache decides cache reuse based on observation-space patch
    similarity alone, risking stale representations at critical moments.
    {\color{blue}Blue}: static tokens, {\color{olive}Yellow}: task-relevant, {\color{red}Red}: overlapping.
    \ourmethod{} additionally inspects the model's own logit margin
    to adaptively invalidate the cache when the model's confidence
    drops.
  }
  \label{fig:teaser}
\end{figure}

We address this gap with \textbf{\ourmethod{}}, a training-free extension
that adds \emph{neural introspection gating}, using the model's own
action-prediction confidence to govern cache
validity (Fig.~\ref{fig:teaser}).

\textbf{logit margin predicts action reliability.}
During autoregressive action decoding, the model produces a softmax
distribution over action tokens at each step.
A large gap between the top-1 and top-2 token probabilities means
the model is confident and cache reuse is safe.
A narrow margin means the model is hesitant between two actions, and
a full recompute is more likely to correct the trajectory.

On the challenging LIBERO-Long suite with OpenVLA, \ourmethod{}
fully recovers the accuracy lost by blind caching (54.8\% vs.\ 50.2\%
for VLA-Cache, 54.0\% for full inference) at a modest compute increase
(1.54 vs.\ 1.43 TFLOPs), remaining 18\% below full inference cost
(1.88 TFLOPs).

\section{Related Work}\label{sec:related}

\textbf{Vision-Language-Action Models.}
VLA~\cite{kim2024openvla,kim2025fine,black2024pi_0,zitkovich2023rt} models unify perception, language understanding, and action prediction
within a single transformer.
RT-2~\cite{zitkovich2023rt} pioneered the approach by fine-tuning a VLM to
output actions as text tokens.
OpenVLA~\cite{kim2024openvla} is a 7B-parameter model built on
Prismatic VLMs~\cite{karamcheti2024prismatic} that performs well on
diverse manipulation benchmarks.
OpenVLA-OFT~\cite{kim2025fine} further improves fine-tuning
efficiency. $\pi_0$~\cite{black2024pi_0} achieves strong results with a flow matching~\cite{lipman2022flow} head.
These models typically adopt large LLM backbones~\cite{touvron2023llama} and are fine-tuned on large robot datasets~\cite{o2024open},
but remain too expensive to run in real time on most robotic hardware.

\textbf{Efficient VLA Inference.}
Several approaches reduce the computational cost of VLA inference.
TinyVLA~\cite{wen2024tinyvla} uses a 422M-parameter model with a diffusion policy~\cite{chi2025diffusion} head for lower latency.
SmolVLA~\cite{shukor2025smolvla} is a 450M-parameter VLA with a flow matching~\cite{lipman2022flow} head.
RoboMamba~\cite{liurobomamba} replaces the Transformer~\cite{vaswani2017attention} backbone with Mamba~\cite{gu2024mamba} for faster inference.
However, these methods require retraining, and their smaller model size limits generalization to new environments. 

\textbf{KV-Cache Optimization for Transformers.}
Key-value caching is standard in autoregressive generation to avoid
redundant computation.
For VLMs and VLAs, several works explore efficient vision-token handling:
FastV~\cite{chen2024fastv} prunes less important vision tokens after a
designated layer;
PyramidDrop~\cite{tang2024pyramiddrop} progressively drops tokens across
layers.
VLA-Cache~\cite{xu2025vlacache} introduces a training-free caching
mechanism that detects unchanged visual patches between frames via cosine
similarity, selectively reuses their KV representations, and employs an
entropy-adaptive layer-wise reuse schedule.
Our work builds directly on VLA-Cache and extends it with a
model-intrinsic gating signal that detects when cached representations
are no longer safe to reuse.

\textbf{Uncertainty Estimation in Neural Networks.}
Using model uncertainty to inform downstream decisions has a long
history~\cite{gal2016dropout,lakshminarayanan2017simple,hendrycks2016baseline}.
Margin-based confidence measures~\cite{settles2009active} are widely used
in active learning and selective prediction.
We adapt this idea to VLA inference, using the logit
margin, available as a zero-cost byproduct of the forward pass, to
gate temporal cache reuse.

\section{Background: VLA-Cache}\label{sec:background}

We briefly review the VLA-Cache~\cite{xu2025vlacache} pipeline upon
which our method builds.

\subsection{VLA Inference}
A VLA model takes as input a camera image $I_t \in \mathbb{R}^{H \times W
\times 3}$ and a language instruction $\ell$, and autoregressively
generates a sequence of $D$ action tokens $a_t = (a_t^1, \dots, a_t^D)$.
The image is divided into $N = (H/p)(W/p)$ non-overlapping patches of
size $p \times p$ (typically $p\!=\!14$, $N\!=\!256$ for $224 \times 224$
images), projected into token embeddings, and concatenated with language
tokens before being fed into the transformer backbone.
The prompt-encoding forward pass through all $L$ layers for all
$N + |\ell|$ input tokens dominates inference time.

\subsection{VLA-Cache Pipeline}
VLA-Cache reduces this cost through three stages:

\textbf{Stage 1: Static Patch Detection.}
Given $I_t$ and $I_{t-1}$, cosine similarity is computed between
corresponding image patches.
Patches with similarity above a threshold $\tau$ (default
0.996) are deemed static and marked as candidates for KV reuse.

\textbf{Stage 2: Task-Relevant Filtering.}
Static patches that receive high text-to-vision attention are removed from
the candidate set, since they are semantically important and should be
freshly computed.
The remaining patches form the reusable set.

\textbf{Stage 3: Entropy-Adaptive Layer-Wise Reuse.}
Different transformer layers exhibit varying attention entropy, which
indicates whether a layer spreads attention broadly or focuses it
narrowly.
VLA-Cache computes per-layer normalized attention entropy and assigns
each layer a different proportion of tokens to reuse: low-entropy layers
reuse more cached tokens, while high-entropy layers recompute more, with
weighted growth smoothing to prevent abrupt transitions.

\section{\ourmethod{}}\label{sec:method}

\begin{figure}[t]
  \centering
  \includegraphics[width=\linewidth]{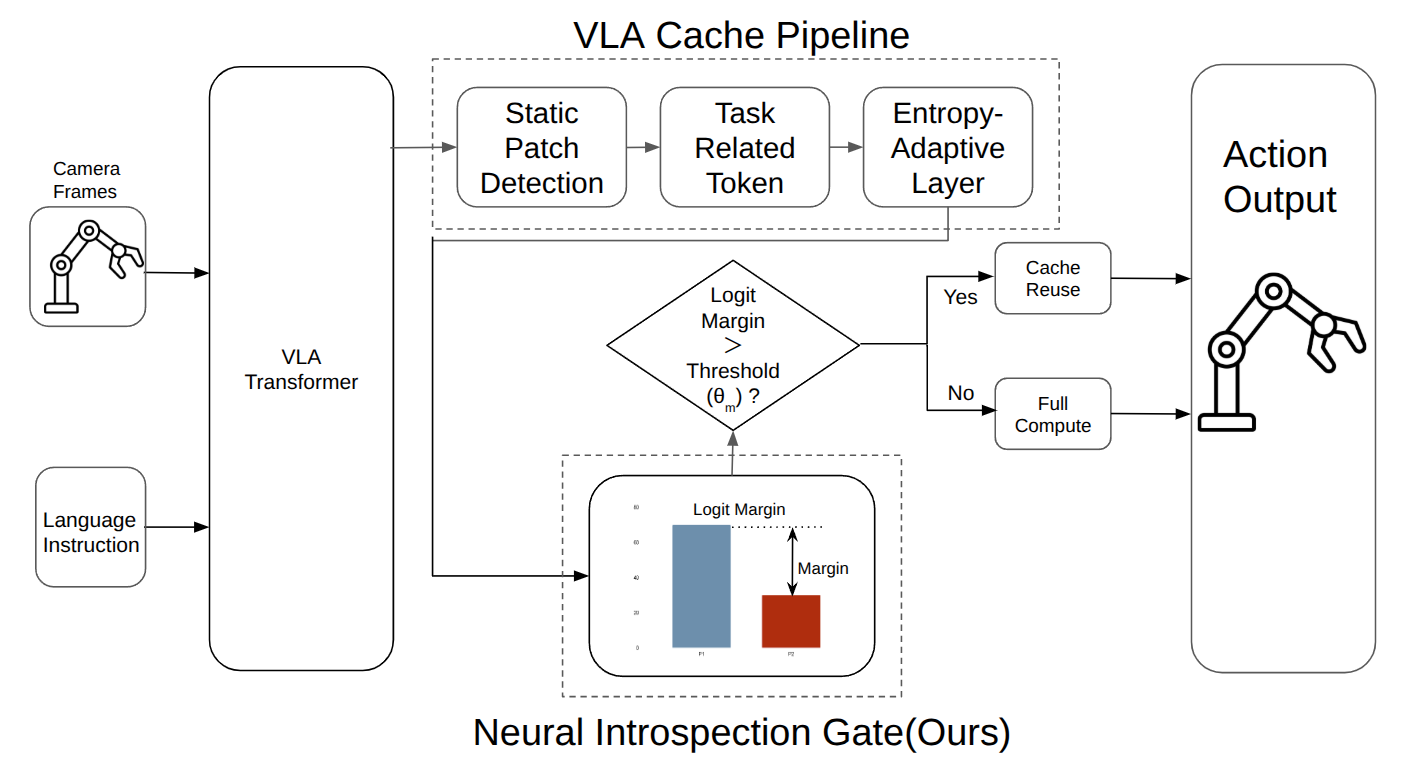}
  \caption{%
    \textbf{\ourmethod{} architecture.}
    At each step $t$, the logit-margin gate evaluates the action
    confidence from step $t{-}1$.
    If the margin signals low confidence, the cache is
    invalidated and full inference runs for step $t$.
    Otherwise, the standard VLA-Cache pipeline proceeds.
  }
  \label{fig:method}
\end{figure}

VLA-Cache's three stages operate entirely in observation space; they
examine what the input looks like but not how the model
responds to it.
\ourmethod{} adds a complementary \emph{gating module} that inspects the
model's own action-prediction confidence to decide whether the cache
should be trusted (Fig.~\ref{fig:method}).

\subsection{Logit-Margin Gating}\label{sec:margin}

During autoregressive decoding of the $D$ action tokens at step $t{-}1$,
the model produces a softmax distribution over the vocabulary for each
action dimension.
We compute the \textbf{mean top-1/top-2 probability margin}:
\begin{equation}\label{eq:margin}
  m_{t-1} = \frac{1}{D} \sum_{d=1}^{D}
    \bigl[ p_1(a_{t-1}^d) - p_2(a_{t-1}^d) \bigr],
\end{equation}
where $p_1$ and $p_2$ denote the highest and second-highest probabilities
in the softmax distribution for each action token.

A large margin indicates a confident prediction; a small margin means
the model is unsure about competing actions, and choosing incorrectly can degrade performance.
Algorithm~\ref{alg:gated} summarizes the full inference procedure.
\begin{figure}[t]
\begin{algorithm}[H]
\caption{\ourmethod{} Inference at Step $t$}
\label{alg:gated}
\begin{algorithmic}[1]
\small
\REQUIRE Image $I_t$, instruction $\ell$, KV cache from step $t{-}1$, \\
    \hspace{1.5em} margin threshold $\theta_m$
\ENSURE Action $a_t$, updated KV cache
\STATE Compute margin $m_{t-1}$ from step $t{-}1$ output scores (Eq.~\ref{eq:margin})
\IF{$m_{t-1} < \theta_m$}
    \STATE Invalidate entire KV cache
    \STATE Run full forward pass on all $N + |\ell|$ tokens
    \STATE Store fresh KV representations in cache
\ELSE
    \STATE Detect static patches via cosine similarity \textit{(Stage 1)}
    \STATE Filter task-relevant patches \textit{(Stage 2)}
    \STATE Apply entropy-adaptive layer-wise reuse \textit{(Stage 3)}
    \STATE Run partial forward pass on non-cached tokens
\ENDIF
\STATE Decode action tokens $a_t = (a_t^1, \dots, a_t^D)$ autoregressively
\RETURN $a_t$, updated cache
\end{algorithmic}
\end{algorithm}
\end{figure}
When the cache is invalidated, the model runs a complete forward pass as
if no cache existed, producing fresh KV representations for all tokens.
These fresh representations then save the cache for subsequent steps.
When the margin is above $\theta_m$, the standard VLA-Cache pipeline
executes with its three-stage reuse mechanism.

\subsection{Properties of the Gating Signal}

\textbf{One-step reactive lag.}
The margin from step $t{-}1$ determines cache policy at step $t$.

\textbf{Self-regulating behavior.}
If cache reuse is safe, the margin stays high and
the gate never fires, so the method reduces to standard VLA-Cache with
negligible overhead.
However, if caching produces uncertain actions, the margin drops below the threshold and triggers a recompute.
The gate activates only when needed, with no prior knowledge of the task difficulty needed.

\begin{figure}[t]
  \centering
  \includegraphics[width=\linewidth]{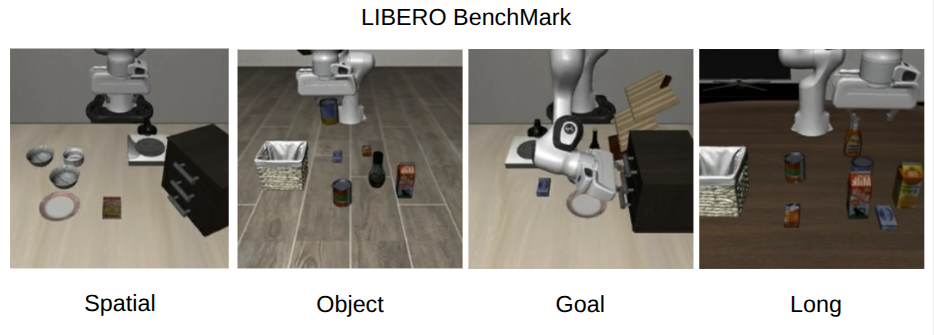}
  \caption{%
    Tasks on LIBERO Benchmark
  }
  \label{fig:libero_benchmark}
\end{figure}
\section{Experiments}\label{sec:experiments}

\subsection{Experimental Setup}

\textbf{Models.}
We evaluate on two VLA architectures:
(a)~OpenVLA~\cite{kim2024openvla}, a 7B-parameter VLA based on
the Prismatic VLM architecture with a Llama-2 backbone, and
(b)~OpenVLA-OFT~\cite{kim2025fine}, which uses optimized
fine-tuning with parallel action-dimension decoding for improved
downstream performance.

\textbf{Benchmark.}
We evaluate on four LIBERO~\cite{liu2023libero} task families:
\textit{Spatial} (10 spatial relationship reasoning tasks),
\textit{Object} (10 object type identification tasks),
\textit{Goal} (10 goal-conditioned tasks), and
\textit{Long} (10 long-horizon multi-step tasks).
Following the standard evaluation protocol, we run 50 episodes per
task (500 episodes per suite).
All experiments are conducted on an NVIDIA A100 GPU. In this research work, we used the UTokyo Azure~\cite{MakotoNakamura20250030}.

\textbf{Baselines.}
We compare:
(a)~\textbf{Full Inference}, the OpenVLA model and OpenVLA-OFT model (8 action chunks for default setting) with official weights for comparability;
(b)~\textbf{VLA-Cache}~\cite{xu2025vlacache}, the observation-heuristic
caching baseline with default hyperparameters ($\tau \!=\!0.996$,
top-$k\!=\!120$).

\textbf{Metrics.}
We report task success rate (\%) and per-step computational cost in
TFLOPs.

\textbf{Hyperparameters.}
The optimal margin scale depends on the specific VLA architecture's
output calibration. We perform a coarse grid search to find a single
threshold $\theta_m$ per architecture that provides the best accuracy in LIBERO-Spatial ($\theta_m = 0.65$ for
OpenVLA; $\theta_m = 0.50$ for OpenVLA-OFT). This single
value is then fixed for all tasks within that specific model architecture.

\subsection{Main Results: OpenVLA}

\begin{figure}[t]
\begin{table}[H]
  \centering
  \caption{%
    \textbf{Task success rates (\%) and compute cost (TFLOPs/step) on
    LIBERO with OpenVLA.}
    \ourmethod{} recovers accuracy lost by VLA-Cache's blind caching
    on tasks where caching causes degradation (Goal, Long), while
    retaining most of the computational savings.
    Best success rates are \textbf{bolded}.
  }
  \label{tab:main_openvla}
  \small
  \begin{tabular}{l cc cc cc}
    \toprule
    & \multicolumn{2}{c}{\textbf{Full}} &
      \multicolumn{2}{c}{\textbf{VLA-Cache}} &
      \multicolumn{2}{c}{\textbf{Ours}} \\
    \cmidrule(lr){2-3}\cmidrule(lr){4-5}\cmidrule(lr){6-7}
    \textbf{Task} & SR\% & TFLOP & SR\% & TFLOP & SR\% & TFLOP \\
    \midrule
    Spatial & 78.8 & 1.89 & 78.8 & 1.43 & \textbf{79.4} & 1.55 \\
    Object  & \textbf{70.4} & 1.86 & 69.4 & 1.44 & 67.8 & 1.54 \\
    Goal    & 77.2 & 1.83 & 74.0 & 1.40 & \textbf{77.4} & 1.50 \\
    Long      & 54.0 & 1.88 & 50.2 & 1.43 & \textbf{54.8} & 1.54 \\
    \midrule
    Avg.    & \textbf{70.1} & 1.87 & 68.1 & 1.43 & 69.9 & 1.53 \\
    \bottomrule
  \end{tabular}
  \vspace{-0.5em}
\end{table}
\end{figure}
Table~\ref{tab:main_openvla} summarizes the results with OpenVLA
across four LIBERO suites (500 episodes each).
The effect of \ourmethod{} depends on how much accuracy VLA-Cache
loses on each suite:

\textbf{Strong recovery on challenging tasks.}
On LIBERO-Goal, VLA-Cache degrades accuracy by 3.2\%
(77.2\%$\rightarrow$74.0\%); \ourmethod{} recovers 106\% of this gap
(77.4\%), slightly exceeding full-inference performance.
On LIBERO-Long, a long-horizon multi-step suite, VLA-Cache drops 3.8\%
(54.0\%$\rightarrow$50.2\%);
\ourmethod{} recovers 121\% of this gap (54.8\%), \emph{exceeding}
full-inference accuracy by 0.8\%.
We suspect this comes from temporal smoothing under selective cache
reuse: on confident steps, cached representations reduce noise from
frame-to-frame variations.

\textbf{Neutral when caching is already safe.}
On LIBERO-Spatial, where VLA-Cache causes no degradation (both at
78.8\%), \ourmethod{} achieves 79.4\% (+0.6\%), since we tuned the hyperparameter $\theta_m$ on this suite.
On LIBERO-Object, where caching causes only a 1.0\% drop, \ourmethod{}
scores 67.8\%, slightly below both baselines.
This likely reflects occasional unnecessary cache invalidation on
tasks with short manipulation horizons where caching is consistently
safe.

\textbf{Average across all suites,}
\ourmethod{} achieves 69.9\% at 1.53~TFLOPs, improving over VLA-Cache
(68.1\%, 1.43~TFLOPs) by 1.8\% while remaining 18\% below full-inference
cost (70.1\%, 1.87~TFLOPs).

\subsection{Main Results: OpenVLA-OFT}
\begin{figure}[t]
\begin{table}[H]
  \centering
  \caption{%
    \textbf{Task success rates (\%) and compute cost (TFLOPs/step) on
    LIBERO with OpenVLA-OFT.}
    OFT achieves high accuracy (93--98\%) across all suites.
    All three methods perform comparably, confirming that \ourmethod{}
    introduces no harm when caching is already safe.
    Best success rates are \textbf{bolded}.
  }
  \label{tab:main_oft}
  \small
  \begin{tabular}{l cc cc cc}
    \toprule
    & \multicolumn{2}{c}{\textbf{Full}} &
      \multicolumn{2}{c}{\textbf{VLA-Cache}} &
      \multicolumn{2}{c}{\textbf{Ours}} \\
    \cmidrule(lr){2-3}\cmidrule(lr){4-5}\cmidrule(lr){6-7}
    \textbf{Task} & SR\% & TFLOP & SR\% & TFLOP & SR\% & TFLOP \\
    \midrule
    Spatial & 93.8 & 4.00 & 93.8 & 3.11 & \textbf{94.0} & 3.23 \\
    Object  & 98.4 & 3.96 & \textbf{98.6} & 3.05 & \textbf{98.6} & 3.08 \\
    Goal    & \textbf{97.2} & 3.94 & 96.8 & 3.06 & 97.0 & 3.07 \\
    Long      & 93.8 & 3.98 & \textbf{94.0} & 3.05 & 93.0 & 3.13 \\
    \midrule
    Avg.    & \textbf{95.8} & 3.97 & \textbf{95.8} & 3.07 & 95.7 & 3.13 \\
    \bottomrule
  \end{tabular}
  \vspace{-0.5em}
\end{table}
\end{figure}
Table~\ref{tab:main_oft} shows results with OpenVLA-OFT.
OFT's improved fine-tuning produces models that are inherently robust to
cache reuse: VLA-Cache causes at most 0.4\% degradation on any suite,
and in some cases (Object, Long) caching slightly improves accuracy
due to its temporal smoothing effect.
Consequently, all three methods achieve nearly identical performance
($\pm$0.8\%), with \ourmethod{} improving slightly over VLA-Cache on
Spatial and Goal, while tying on Object and dropping slightly on Long.

This result supports the \emph{self-regulating} behavior of \ourmethod{}: when
cache reuse is already safe (as with OFT), the margin remains high,
the gate rarely triggers, and the method adds negligible overhead.
In other words, the gate is most useful where OpenVLA struggles
under blind caching and mostly dormant otherwise.

\subsection{Accuracy Compute Trade-off}

\begin{figure}[t]
  \centering
  \includegraphics[width=\linewidth]{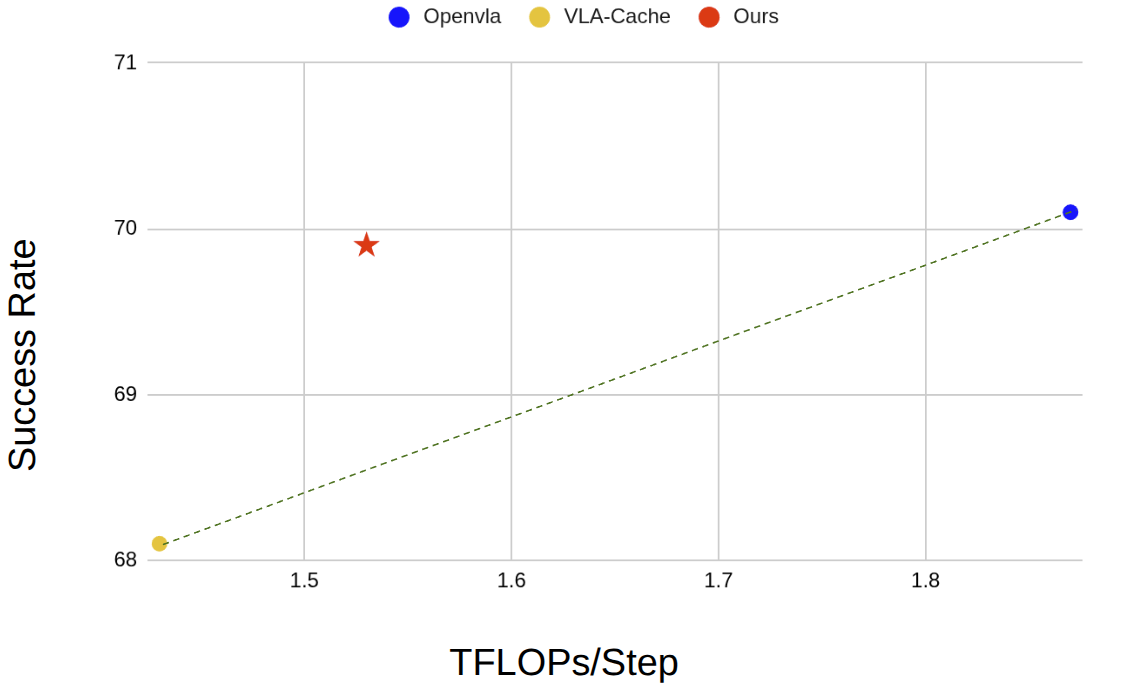}
  \caption{%
    \textbf{Accuracy compute trade off} on average score of four LIBERO Benchmark(OpenVLA).
    \ourmethod{} (star) lies above the line connecting Full Inference and
    VLA-Cache, indicating Pareto-optimality: it achieves full-inference
    accuracy at substantially lower compute.
  }
  \label{fig:pareto}
\end{figure}

Fig.~\ref{fig:pareto} plots the accuracy compute trade off on the
average score of LIBERO Benchmark in OpenVLA. The VLA-Cache causes the most degradation but the least computation. \ourmethod{} achieves 69.9\% at 1.53~TFLOPs, at 18\% below full-inference cost (1.87~TFLOPs) while dropping only 0.2\% accuracy. \ourmethod{} lies above the linear interpolation between
VLA-Cache and Full Inference, making it Pareto-optimal.
\subsection{When Does Gating Fire?}

\begin{figure}[t]
  \centering
  \includegraphics[width=\linewidth]{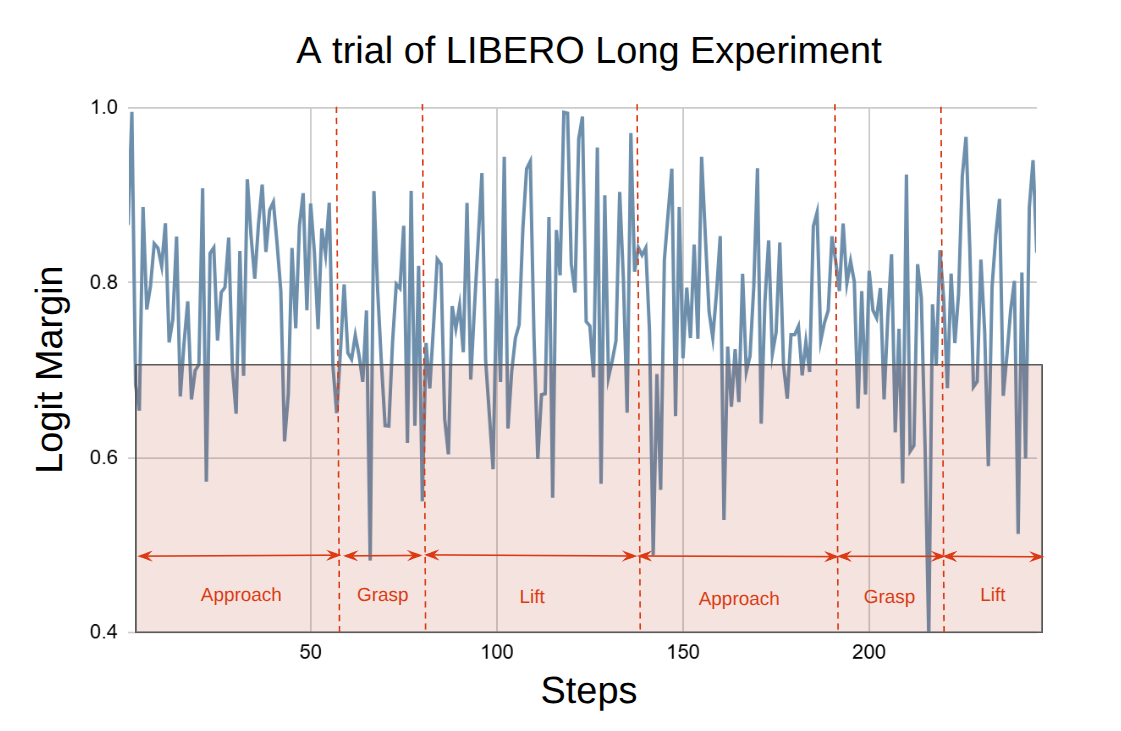}
  \caption{%
    \textbf{Gating trace} for a representative LIBERO-Long episode.
    The logit margin (blue) dips during critical manipulation phases
    (grasping objects). Pink regions mark gating events. Each manipulation has three phases: approach, grasp, and lift.
  }
  \label{fig:gating_trace}
\end{figure}

Fig.~\ref{fig:gating_trace} visualizes the logit-margin signal during a
representative LIBERO-Long episode.
The margin shows a clear temporal pattern: it stays high during
segments where the robot is executing a straightforward motion
(e.g., approaching an object), and dips during critical manipulation
phases where the action space is genuinely ambiguous (e.g., grasp
alignment, transition between sub-tasks).
Cache invalidation events cluster around these critical
moments.
The gate fires where fresh representations matter most.

On average across LIBERO-Long episodes, the gate fires on approximately
24\% of steps: 
\ourmethod{} reuses cached representations 76\% of the time, triggering
full recompute only when the model's confidence drops.

\subsection{Threshold Sensitivity}

\begin{figure}[t]
  \centering
   \includegraphics[width=\linewidth]{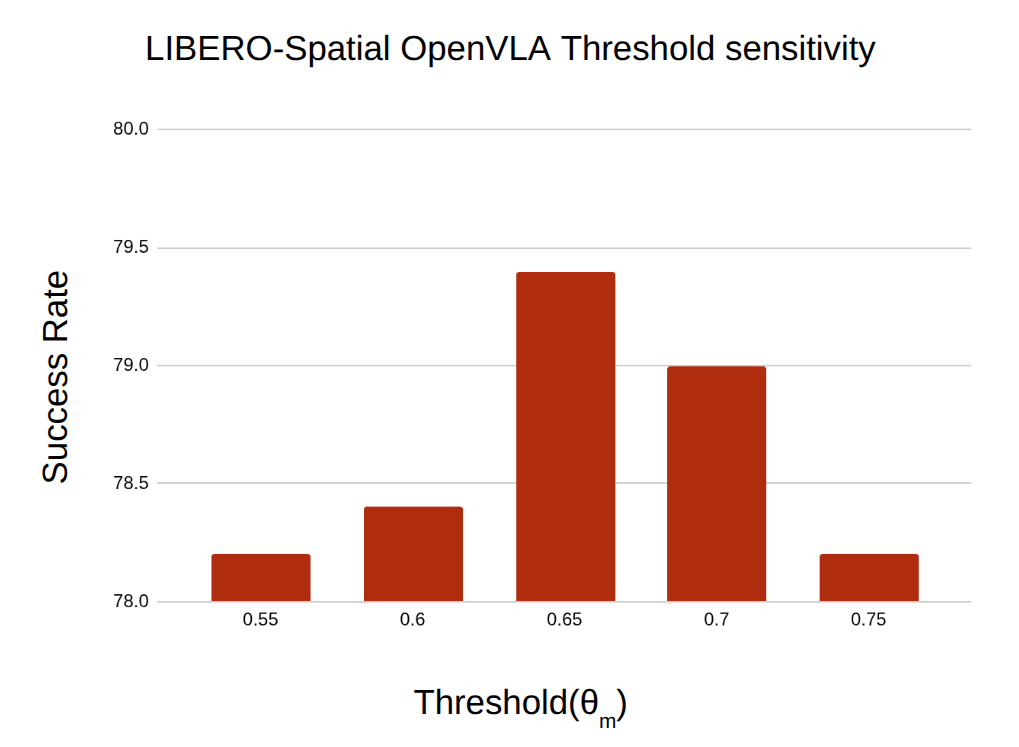}
  \caption{%
    \textbf{Threshold sensitivity} on LIBERO-Spatial (OpenVLA).
    The method exhibits a broad plateau of near-optimal performance.
  }
  \label{fig:vla_sensitivity}
\end{figure}

\begin{figure}[t]
  \centering
   \includegraphics[width=\linewidth]{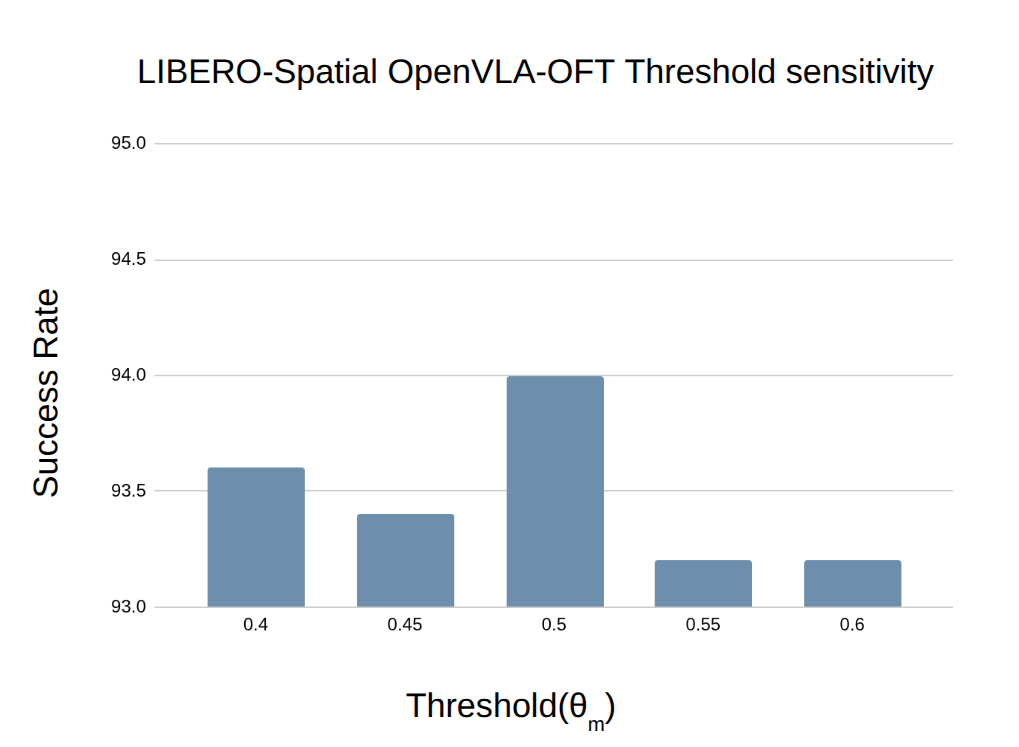}
  \caption{%
    \textbf{Threshold sensitivity} on LIBERO-Spatial (OpenVLA-OFT).
    The method exhibits a broad plateau of near-optimal performance.
  }
  \label{fig:vla_oft_sensitivity}
\end{figure}
Fig.~\ref{fig:vla_sensitivity} shows the effect of varying the margin
threshold $\theta_m$.
Success rate is stable across a broad range: sweeping $\theta_m$ from 0.55 to 0.75
on LIBERO-Spatial yields success rates tightly bounded between 78.2\%
and 79.4\%.
At very low thresholds ($\theta_m \rightarrow 0$), the gate never fires
and the method reduces to VLA-Cache. Conversely, at very high
thresholds, the gate fires too often, causing unnecessary
invalidation that can disrupt the temporal smoothing benefits of caching
and slightly underperform the baseline (e.g., 78.2\% at $\theta_m = 0.75$).
The chosen threshold ($\theta_m = 0.65$) sits in a stable region
that balances accuracy recovery and compute savings.

This robustness transfers to the OpenVLA-OFT
architecture as well (Fig.~\ref{fig:vla_oft_sensitivity}). While the calibrated center point
shifts due to the different action head formulation ($\theta_m = 0.5$
for OFT), the stability remains: sweeping $\theta_m$ from 0.40 to 0.60
on LIBERO-Spatial yields success rates tightly bounded between 93.2\%
and 94.0\%.
The gating module does not need precise hyperparameter tuning.

\section{Discussion}\label{sec:discussion}

\textbf{When does gating help?}
The pattern is simple: \ourmethod{} helps most on tasks where
VLA-Cache causes clear accuracy degradation.
On LIBERO-Goal ($-3.2$\%) and LIBERO-Long ($-3.8$\%), where cache
staleness compounds over long manipulation sequences, our gating
recovers over 100\% of the lost accuracy (106\% and 121\%, respectively).
On suites where caching is already safe (Spatial, Object), the gating
rarely triggers and adds minimal overhead.
In practice, this means users do not need to predict in advance which
tasks will be sensitive to cache reuse. \ourmethod{} activates when needed.

\textbf{Why logit margin?}
The logit margin captures the model's output uncertainty:
its difficulty in selecting a single best action from the vocabulary.
This is more direct than
observation-space signals, which only measure what has changed
visually without knowing how those changes affect predictions.
A visually minor change (e.g., a subtle rotation of a thin object) can
dramatically alter the model's action prediction but go undetected by
pixel similarity.
The margin reflects this internal effect directly.

In our initial experiments, we evaluated alternative uncertainty signals, including the entropy of the action distribution and attention entropy, but found both ineffective for temporal gating. We suspect that action entropy fails because the model assigns near-zero probability to the vast majority of its large vocabulary (e.g., 32,000 tokens for OpenVLA). In this regime, the mathematical difference in total entropy between a confident prediction (e.g., a 90\% vs.\ 5\% split for the top choices) and an unconfident prediction (e.g., a 45\% vs.\ 40\% split) is overwhelmed by the static noise of the long-tail vocabulary.

Similarly, attention entropy fails as a temporal gating signal due to the attention sink phenomenon~\cite{xiao2023efficient} pervasive in deep autoregressive language models. In LLM-based architectures like OpenVLA, the model structurally allocates a massive proportion of attention mass to initial syntactic tokens. Because these sink tokens absorb the majority of the probability mass, the overall entropy of the attention distribution remains artificially anchored throughout an episode. Consequently, meaningful dynamic shifts in visual attention (e.g., shifting focus from the gripper to the target object) barely perturb the total entropy.

We also experimented with the action delta between consecutive steps as a gating signal. However, we found that it primarily correlates with manipulation phase transitions rather than model uncertainty: action deltas are large during approach and grasp phases, and small during post-grasp lifting. This makes it a phase-change indicator, not an uncertainty signal, sharing the same fundamental blind spot as observation-space heuristics.

The top-1/top-2 logit margin avoids both problems. By strictly isolating the probability gap between the two most likely action tokens, it ignores vocabulary noise entirely. It explicitly measures the model's internal conflict at the bottleneck of decision-making, triggering a recomputation only when the cached representation lacks the discriminatory power to resolve the action safely. 

\textbf{Self-regulating compute.}
The main advantage of margin-based gating is that the invalidation
frequency naturally adapts to task difficulty.
On simple tasks (Spatial, Object), the model is consistently confident
and the gate rarely fires ($<$10\% of steps), preserving nearly all of
VLA-Cache's speedup.
On hard tasks (Goal, Long), the model is frequently uncertain
($\sim$24\% of steps), triggering more recomputation where it is most
needed.
No task-specific tuning is required.

\textbf{Cross-architecture generality.}
We evaluate on two VLA architectures with different
action prediction strategies: OpenVLA uses discrete token
decoding while OpenVLA-OFT uses parallel action-dimension decoding.
The gating signal transfers without modification because it is
extracted from the shared transformer backbone's output logits,
which are present in both architectures.

\textbf{Limitations.}
(1)~The logit margin cannot distinguish between cache-induced
uncertainty and inherent model ambiguity.
In practice, unnecessary recomputation at difficult moments
incurs a minor compute overhead but no accuracy penalty. Invalidation
is always safe, just occasionally wasteful.
(2)~The margin is computed from step $t{-}1$ and used to
decide cache reuse at step $t$, introducing a one-step reactive lag.
(3)~The method introduces one hyperparameter ($\theta_m$).
While we show it generalizes zero-shot across distinct task suites
for a given model, although the threshold is robust (Figs.~\ref{fig:vla_sensitivity} and~\ref{fig:vla_oft_sensitivity}), its optimal value depends on the specific VLA
architecture's output calibration. Thus, $\theta_m$ must be calibrated
once per model family (e.g., via grid search on a small validation set).
Automatic threshold adaptation (e.g., via online margin statistics)
could remove this requirement in future work.
(4)~We evaluate in simulation. Real robot validation remains future work.

\section{Conclusion}\label{sec:conclusion}

We presented \ourmethod{}, a lightweight, training-free method for
improving KV-cache reuse in Vision-Language-Action models.
By monitoring the model's logit margin, the probability gap between the
top-two predicted action tokens, \ourmethod{} decides when cached
representations are safe to reuse and when a full recompute is needed.
On LIBERO with OpenVLA, our method recovers over 100\% of the
accuracy lost by blind caching on challenging long-horizon and
goal-conditioned tasks, while retaining 80\% of the computational
savings. On OpenVLA-OFT, where caching is already safe, the gate remains
dormant and introduces no degradation.
The method requires no additional parameters, no retraining, and can
be integrated transparently into any autoregressive VLA inference stack.

\balance

\bibliographystyle{ieeetr}

\bibliography{refs}

\end{document}